\documentclass[11pt, a4paper, logo, twocolumn, copyright]{googledeepmind}

\pdftrailerid{redacted}

\makeatletter
\renewcommand\bibentry[1]{\nocite{#1}{\frenchspacing\@nameuse{BR@r@#1\@extra@b@citeb}}}
\makeatother

\usepackage{kantlipsum, lipsum}
\usepackage{dsfont}
\usepackage{dm-colors}

\usepackage{algorithm}
\usepackage{algpseudocode}
\usepackage{adjustbox}
\usepackage{multirow}
\usepackage{nicefrac}
\algrenewcommand\algorithmicindent{0.7em}

\algblock{ParFor}{EndParFor}
\algnewcommand\algorithmicparfor{\textbf{parallel for}}
\algnewcommand\algorithmicpardo{\textbf{do}}
\algnewcommand\algorithmicendparfor{\textbf{end\ parallel for}}
\algrenewtext{ParFor}[1]{\algorithmicparfor\ #1\ \algorithmicpardo}
\algrenewtext{EndParFor}{\algorithmicendparfor}

\newcolumntype{R}[2]{%
    >{\adjustbox{angle=#1,lap=\width-(#2)}\bgroup}%
    l%
    <{\egroup}%
}

\usepackage[authoryear, sort&compress, round]{natbib} 
\usepackage{comment}

\usepackage{pifont}

\usepackage{subcaption}
\usepackage{url}

\graphicspath{{figures/}}

\title{Eager Updates For Overlapped Communication and Computation in DiLoCo}

\correspondingauthor{douillard@google.com}

\keywords{eager updates, distributed learning, large-scale} 

\reportnumber{} 

\renewcommand{\today}{} 

\author[*,$\dagger$,2]{Satyen Kale}
\author[*,1]{Arthur Douillard}
\author[1]{Yanislav Donchev}

\affil[1]{Google DeepMind}
\affil[2]{Google Research}
\affil[*]{Equal core contributions}
\affil[$\dagger$]{Currently at Apple.}

\begin{abstract}
Distributed optimization methods such as DiLoCo have been shown to be effective in training very large models across multiple distributed workers, such as datacenters. These methods split updates into two parts: an \emph{inner optimization} phase, where the workers independently execute multiple optimization steps on their own local data, and an \emph{outer optimization} step, where the inner updates are synchronized. While such approaches require orders of magnitude less communication than standard data-parallel training, in settings where the workers are datacenters, even the limited communication requirements of these approaches can still cause significant slow downs due to the blocking necessary at each outer optimization step. In this paper, we investigate techniques to mitigate this issue by overlapping communication with computation in a manner that allows the outer optimization step to fully overlap with the inner optimization phase. We show that a particular variant, dubbed \emph{eager} updates, provides competitive performance with standard DiLoCo in settings with low bandwidth between workers.
\end{abstract}

\begin{document}

\maketitle

\section{Introduction} \label{sec:intro}

As language models and data sets get ever larger, it has become increasingly important to resort to distributed training approaches to effectively handle the larger scales. A particularly effective technique, DiLoCo, was proposed by \citet{douillard2023diloco}. DiLoCo leverages techniques from Federated Learning~\citep{mcmahan2017fedavg} and uses a particular instantiation similar to the FedOpt algorithm \citep{reddi2021adaptive}. 

\begin{figure*}[ht!]
\centering
    \includegraphics[width=1\linewidth]{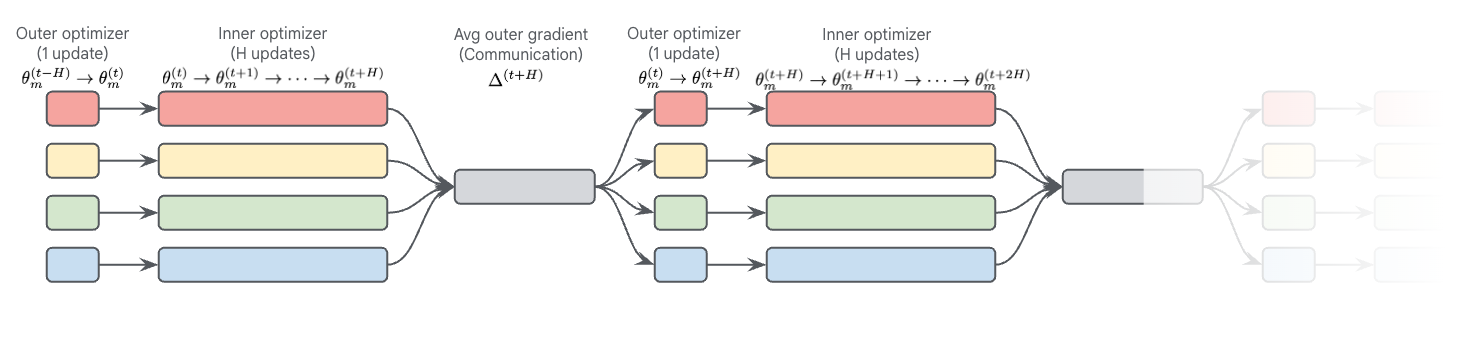}
    \caption{\textbf{Data flow and operations in standard DiLoCo.} Here, 4 workers execute in parallel and alternate sequentially computation (the outer and inner optimization steps) and communication (averaging outer gradients across workers).}
\label{fig:std-diloco}
\end{figure*}

Specifically, training is split into inner and outer optimization phases. In each inner optimization phase, all workers independently execute an optimizer (typically, AdamW) on their local data starting from the current values of the global parameters. Then, all workers communicate their updates to each other in the outer optimization phase. These updates are aggregated via an all-reduce operation into a single ``outer gradient'', which is then applied via an \emph{outer optimizer} (typically, Nesterov Momentum) to the current global parameters to get the new global parameters, which then form the starting point for the next inner optimization phase. 
The pseudocode appears in \autoref{alg:std-diloco} and visualization in \autoref{fig:std-diloco}.

This approach compares favorably with standard data-parallel distributed training (in which each worker computes one gradient on a batch of local data, after which all gradients are aggregated into one and applied to the current parameters). See \citep{douillard2023diloco} for details. The benefit is that the total communication requirements goes down by a factor of the number of inner optimization steps (typically, 50-100) compared to standard data-parallel, leading to better running time.

However, in certain settings such as cross-datacenter training, communication links between workers have low bandwidth, and workers are forced to block in each outer optimization phase until all the outer gradients are communicated before continuing computation. This leads to wasted time due to idle compute. 

The goal of this paper is to mitigate this issue by developing techniques to overlap communication with computation leading to better compute utilization. The particular approach studied here is to allow the communication of outer gradients to happen in parallel with the computation of the immediate next inner optimization phase. I.e., at the end of each inner optimization phase, workers dispatch the computed outer gradients to be communicated to the other workers, and then \emph{immediately} start executing the next inner optimization phase without waiting for the outer gradient all-reduce to finish. The all-reduce operation is allowed as much as time as it takes for the inner optimization phase to complete. Thus, at the end of the inner optimizaiton phase, the all-reduced outer gradient from the previous inner optimization phase is available, and can be applied to the local parameters. 

A na\"ive implementation of the above approach leads to worse convergence than standard DiLoCo since the all-reduced outer gradients are applied with a delay of one entire inner optimization phase. To improve on this, we develop an \emph{eager} version of this approach based on the following idea. Note that the \emph{local} outer gradient computed at each worker is already available to that worker before the all-reduce with the other, non-local, outer-gradients. This local outer gradient can serve as a good proxy for the all-reduced outer-gradients, and can be used in an outer optimization step at each worker before starting the next inner optimization phase. Then at the end of that inner optimization step, when the all-reduced outer gradients become available to the worker, the (delayed) non-local outer gradients can be applied to the parameters, along with the new (fresh) local outer gradient. We call this method \emph{eager} since it eagerly applies local gradients without waiting for the non-local gradients to arrive at the worker.

Our experiments show that eager updates significantly help reduce the performance hit of na\"ive delayed outer gradients and achieve training loss close to standard DiLoCo. When factoring in \emph{compute utilization} however, eager updates significantly improve over standard DiLoCo. We provide details in the following sections.

\section{Algorithms} \label{sec:algorithms}
\begin{figure*}[ht!]
\centering
    \includegraphics[width=1\linewidth]{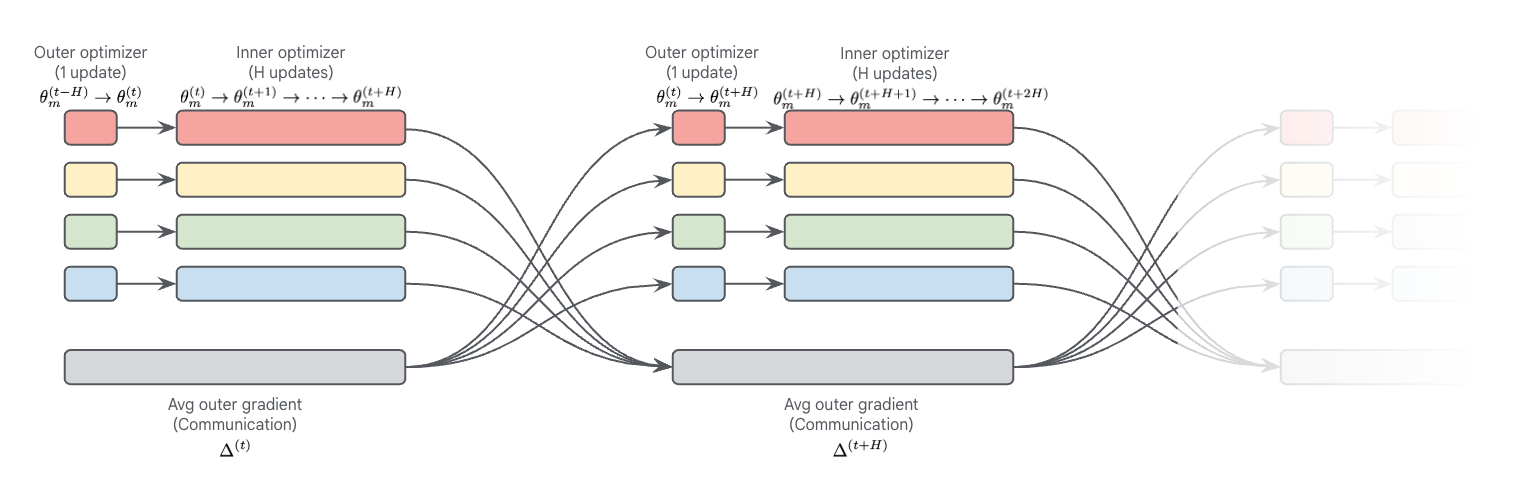}
    \caption{\textbf{Data flow and operations in DiLoCo with delayed outer gradients.} Here, 4 workers execute optimization steps in parallel with each other, as well as with the communication required for averaging outer gradients. This is accomplished by delaying the application of the averaged outer gradient in the outer optimizer.}
\label{fig:delayed-diloco}
\end{figure*} 
In this section we describe the algorithms studied in detail. For all algorithms, we denote the model parameters as $\theta$. We use the superscript notation $\theta^{(t)}$ to indicate the parameters at a given step $t$, and the subscript notation $\theta_m$ to denote a particular shard of the DiLoCo replica. For example, $\theta^{(t)}_m$ indicates the parameters of DiLoCo replica $m$ at step $t$. If no subscript is used, the parameters are replicated across DiLoCo replicas. Note that it is possible for parameters to not be replicated and yet to be of the same value. 

\subsection{Standard DiLoCo} \label{sec:model_diloco}
DiLoCo is an instantiation of the FedOpt framework of \citet{reddi2021adaptive} applied to language models which is a bi-level federated optimization paradigm using an \emph{inner optimizer}, Adam \citep{kingma2014adam}, and an \emph{outer optimizer}, SGD with Nesterov momentum \citep{sutskever2013nesterov}. The DiLoCo algorithm is shown in \autoref{alg:std-diloco} and visualized in \autoref{fig:std-diloco}.

\begin{algorithm}[!h]
\caption{DiLoCo} \label{alg:std-diloco}
\begin{algorithmic}[1]
\Require $M$ replicas
\Require Synchronization frequency $H$
\Require Model replicas $\{\theta^{(0)}_1, \dots, \theta^{(0)}_M\}$
\Require Data shards $\{\mathcal{D}_1, \dots, \mathcal{D}_M\}$
\Require Optimizers $\texttt{InnerOpt}$ and $\texttt{OuterOpt}$
\ParFor{\texttt{replica $m = 1 \ldots M$}} 
\For{\texttt{step $t = 1 \ldots T$}}
    \State $x \sim \mathcal{D}_m$
    \State $\mathcal{L} \gets f(x, \theta_m^{(t-1)})$
    \State $\theta_m^{(t)} \gets \texttt{InnerOpt}(\theta_m^{(t-1)}, \nabla_\mathcal{L})$
    \item[]
    \If{$t\mod H == 0$}
        \State $\Delta^{(t)}_{m} \gets \theta^{(t-H)}_{m} - \theta_{m}^{(t)}$
        \State $\Delta^{(t)} \gets {\small \texttt{async-send}}[\frac{1}{M} \sum_{m=1}^M (\Delta^{(t)}_{m})]$
        \State $\texttt{block-receive}[{\Delta^{(t)}}]$
        \State $\theta_m^{(t)} \gets \texttt{OuterOpt}(\theta_m^{(t-H)}, \Delta^{(t)})$
    \EndIf
\EndFor
\EndParFor
\end{algorithmic}
\end{algorithm}

In DiLoCo, $M$ local replicas perform, in parallel, $H$ steps of the inner optimizer \texttt{InnerOpt} on a different subsets of the data (L3 to L5 in \autoref{alg:std-diloco}). Every $H$ steps, each replica computes an \textit{outer gradient} $\Delta_m^{(t)} = \theta_m^{(t-H)} - \theta_m^{(t)}$ (L7), a delta in the parameter space, and communicates it to all other replicas. This communication can be performed through a central parameter server or through direct communication of each worker to the others (e.g. with a ring all-reduce), and results in each worker obtaining $\Delta^{(t)} = \nicefrac{1}{M} \sum_{m=1}^M \Delta^{(t)}_m$ (L7-9). This outer gradient is applied to the \textit{outer parameters}, which are the previously synchronized parameters $\theta_m^{(t-H)}$, using the outer optimizer \texttt{OuterOpt} (L10). 

The costly communication between non-colocated devices happens during the averaging of outer gradients, in lines 8 and 9 of \autoref{alg:std-diloco}. While the communication cost at each outer optimization step is exactly the same as in each iteration of standard Data-Parallel training, since it is done every $H$ (e.g., one hundred) steps, the communication cost is amortized.


DiLoCo is a successful instantiation of FedOpt applied to language models where the inner optimizer is Adam \citep{kingma2014adam} and the outer optimizer is SGD with Nesterov momentum \citep{sutskever2013nesterov}. 

\subsection{Na\"ive Delayed Outer Gradients} 
\autoref{alg:na\"ive-delayed} gives the pseudocode for na\"ive delayed outer gradients in DiLoCo. A visualization is given in \autoref{fig:delayed-diloco} which indicates how delaying the application of the outer gradients in the outer optimizer enables effective overlapping of communication with computation.
\begin{algorithm}[!h]
\caption{Na\"ive Delayed Outer Gradients in DiLoCo} \label{alg:na\"ive-delayed}
\begin{algorithmic}[1]
\Require $M$ replicas
\Require Synchronization frequency $H$
\Require Model replicas $\{\theta^{(0)}_1, \dots, \theta^{(0)}_M\}$
\Require Data shards $\{\mathcal{D}_1, \dots, \mathcal{D}_M\}$
\Require Optimizers $\texttt{InnerOpt}$ and $\texttt{OuterOpt}$
\ParFor{\texttt{replica $m = 1 \ldots M$}} 
\For{\texttt{step $t = 1 \ldots T$}}
    \State $x \sim \mathcal{D}_m$
    \State $\mathcal{L} \gets f(x, \theta_m^{(t-1)})$
    \State $\theta_m^{(t)} \gets \texttt{InnerOpt}(\theta_m^{(t-1)}, \nabla_\mathcal{L})$
    \item[]
    \If{$t\mod H == 0$}
        \State $\Delta^{(t)}_{m} \gets \theta^{(t-H)}_{m} - \theta_{m}^{(t)}$
        \State $\Delta^{(t)} \gets {\small \texttt{async-send}}[\frac{1}{M} \sum_{m=1}^M (\Delta^{(t)}_{m})]$
        \If{$t > H$}
            \State $\texttt{block-receive}[{\Delta^{(t-H)}}]$
            \State $\theta_m^{(t)} \gets \texttt{OuterOpt}(\theta_m^{(t-H)}, \Delta^{(t-H)})$
        \EndIf
    \EndIf
\EndFor
\EndParFor
\end{algorithmic}
\end{algorithm}

The main differences from standard DiLoCo to note are: 
\begin{enumerate}
    \item Each worker maintains its own copy of the model parameters $\theta_i^{(t)}$, which is never synchronized across workers. Thus, the model parameters at the workers may diverge from each other, and may benefit from periodic synchronizing by simple averaging. In experiments, however, we found no benefit to this period synchronization.
    \item In the outer optimization step (L10--11), outer gradients from the \emph{previous} inner optimization phase are used instead of the current ones (i.e. $\Delta^{(t-H)}$ instead of $\Delta^{(t)}$. Effectively, this means that the \texttt{async-send} operation in L8 can be executed in parallel with the next inner optimization phase, since its result is only consumed at the end of that phase.
\end{enumerate}

\subsection{Eager updates with delayed outer gradients} 

\begin{algorithm}[!h]
\caption{Eager Updates with Delayed Outer Gradients in DiLoCo} \label{alg:eager-delayed}
\begin{algorithmic}[1]
\Require $M$ replicas
\Require Synchronization frequency $H$
\Require Model replicas $\{\theta^{(0)}_1, \dots, \theta^{(0)}_M\}$
\Require Data shards $\{\mathcal{D}_1, \dots, \mathcal{D}_M\}$
\Require Optimizers $\texttt{InnerOpt}$ and $\texttt{OuterOpt}$
\ParFor{\texttt{replica $m = 1 \ldots M$}} 
\For{\texttt{step $t = 1 \ldots T$}}
    \State $x \sim \mathcal{D}_m$
    \State $\mathcal{L} \gets f(x, \theta_m^{(t-1)})$
    \State $\theta_m^{(t)} \gets \texttt{InnerOpt}(\theta_m^{(t-1)}, \nabla_\mathcal{L})$
    \item[]
    \If{$t\mod H == 0$}
        \State $\Delta^{(t)}_{m} \gets \theta^{(t-H)}_{m} - \theta_{m}^{(t)}$
        \State $\Delta^{(t)} \gets {\small \texttt{async-send}}[\frac{1}{M} \sum_{m=1}^M (\Delta^{(t)}_{m})]$
        \If{$t > H$}
            \State $\texttt{block-receive}[{\Delta^{(t-H)}}]$
            \State $\tilde{\Delta}_m^{(t)} \gets \frac{1}{M} (\Delta_m^{(t)} - \Delta_m^{(t-H)}) + \Delta^{(t-H)}$.
            \State $\theta_m^{(t)} \gets \texttt{OuterOpt}(\theta_m^{(t-H)}, \tilde{\Delta}_m^{(t)})$
        \EndIf
    \EndIf
\EndFor
\EndParFor
\end{algorithmic}
\end{algorithm}

\autoref{alg:na\"ive-delayed} gives the pseudocode for eager updates with delayed outer gradients in DiLoCo. The main difference from \autoref{alg:na\"ive-delayed}, na\"ive delayed outer gradients, is in lines 11 and 12. Line 11 computes a ``fresher'' version of the delayed outer gradient by adding the current local outer gradient and removing the stale local outer gradient, both appropriately scaled. In other words, the computation in line 11 can be equivalently written as $\tilde{\theta}_m^{(t)} \gets \frac{1}{M}(\Delta_m^{(t)} + \sum_{m' \neq m} \Delta_{m'}^{(t-H)})$, which brings out the fact that we're just computing an average of the current local outer gradient and all the stale non-local outer gradients. Crucially, just like in the na\"ive implementation, this computation only requires outer gradients from the \emph{previous} inner optimization phase are used instead of the current ones (i.e. $\Delta^{(t-H)}$ instead of $\Delta^{(t)}$), and so the \texttt{async-send} in line 8 can be executed in parallel with the next inner optimization phase.

\begin{figure*}[t]
\centering
\captionsetup[subfigure]{justification=centering}
\begin{subfigure}{0.325\linewidth}
  \centering
  \includegraphics[width=1\linewidth]{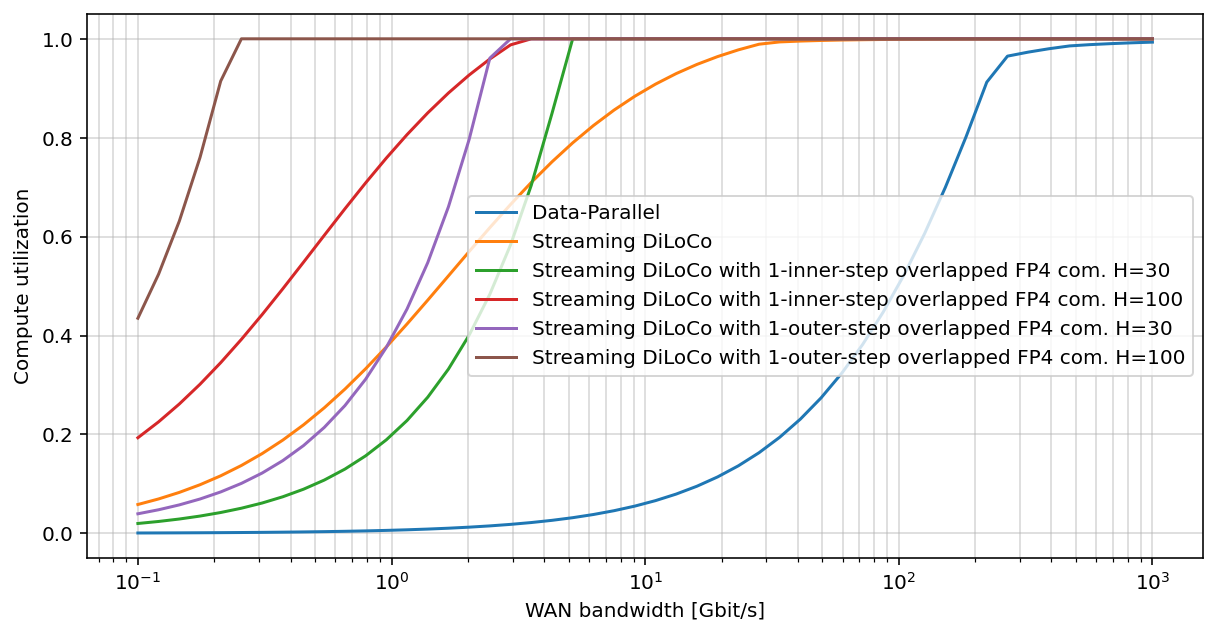}
  \caption{1B parameters model.}
  \label{fig:bandwdith_1b}
\end{subfigure}\hfill
\begin{subfigure}{0.325\linewidth}
  \centering
  \includegraphics[width=1\linewidth]{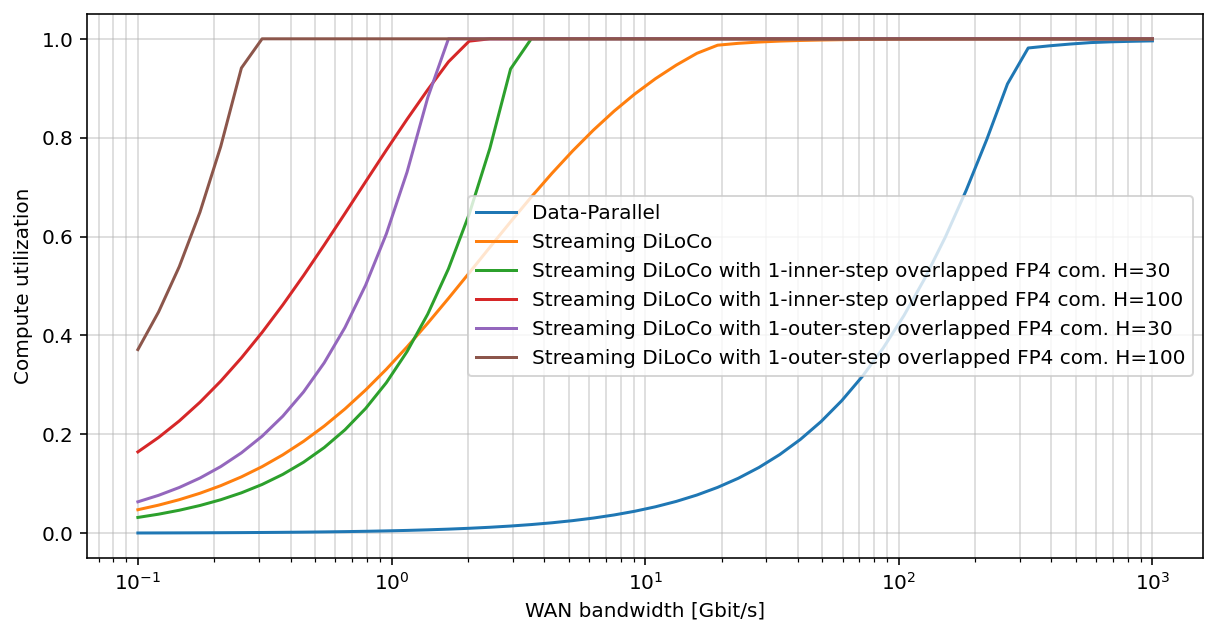}
  \caption{10B parameters model}
  \label{fig:bandwdith_1b}
\end{subfigure}\hfill
\begin{subfigure}{0.325\linewidth}
  \centering
  \includegraphics[width=1\linewidth]{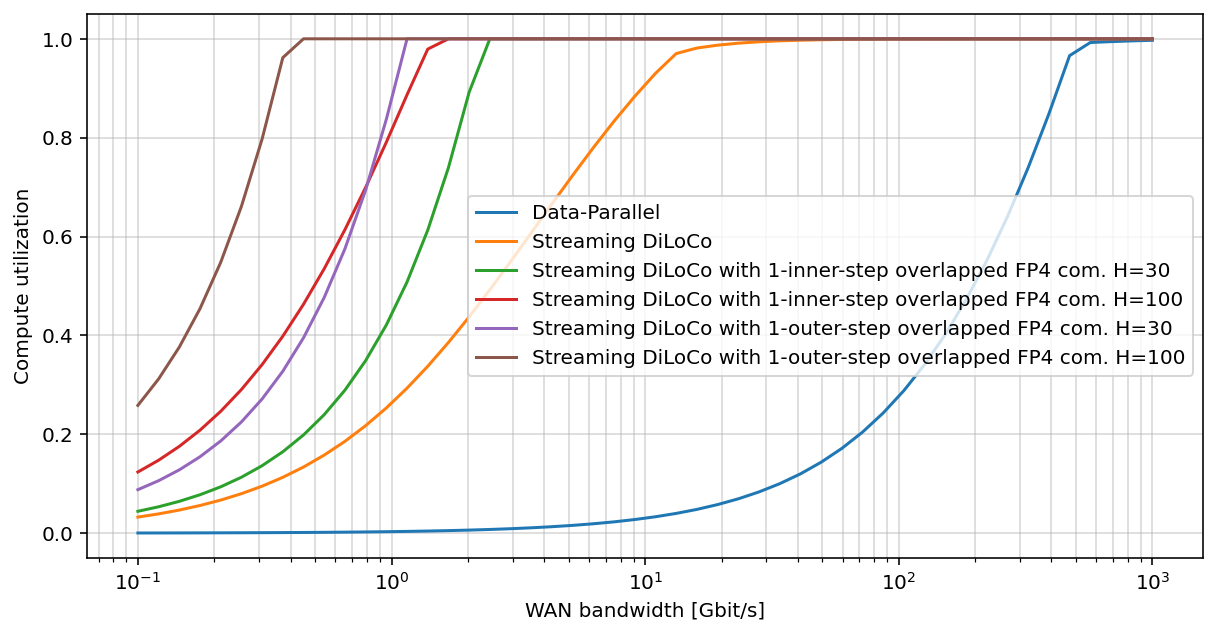}
  \caption{100B parameters model}
  \label{fig:bandwdith_100b}
\end{subfigure}
\caption{\textbf{Compute Utilization} simulated across a range of bandwidth. A compute utilization of 0.8 means 80\% of the time is spent in computation, and 20\% in communication. Our best method reaches a compute utilization of 95\% for models 1B, 10B, and 100B with a bandwidth roughly constant between 1 and 5 Gbit/s. Data-Parallel on the other hand requires 100, 200, and 300Gbit/s.}
\label{fig:bandwdith}
\end{figure*}

\section{Experiments}\label{sec:experiments}

We perform our experiments with a Chinchilla architecture \citep{hoffmann2022chinchilla}. Following \cite{wortsman2023smallscaleproxieslargescaletransformer} and \cite{jaghouar2024intellect1}, we use QKNorm \citep{henry2020querykeynormalization} and a Z-loss \citep{chowdhery2023palm} with a factor of 1e-4 to stabilize training. We report in \autoref{tab:hp_architecture} the architecture hyperparameters and token budget at each scale. Unlike the recommendation in Post-Local SGD \citep{Lin2020_localsgd}, we train all our models from scratch. The main hyperparameter of DiLoCo is its outer learning rate; we tuned it to be optimal at small scale at $0.4$, and kept it fixed across all scales. For all experiments we use Streaming DiLoCo \citep{douillard2025streamingdiloco} instead of standard DiLoCo \citep{douillard2023diloco}. Streaming DiLoCo essentially applies standard DiLoCo to different parts of the models on different schedules. While we presented our delayed outer gradients methods only for standard DiLoCo, it can be easily applied to streaming DiLoCo.

We use the C4 dataset \citep{c4} and train models from 35 million to 1 billion parameters. Each scale is trained with the Chinchilla-optimal number of steps. We use 2 DiLoCo replicas, each of them performing FSDP \citep{zhao2023fsdp} across their respective closely located devices.

\begin{figure*}[t]
\centering
\captionsetup[subfigure]{justification=centering}
\begin{subfigure}{0.45\linewidth}
  \centering
  \includegraphics[width=1\linewidth]{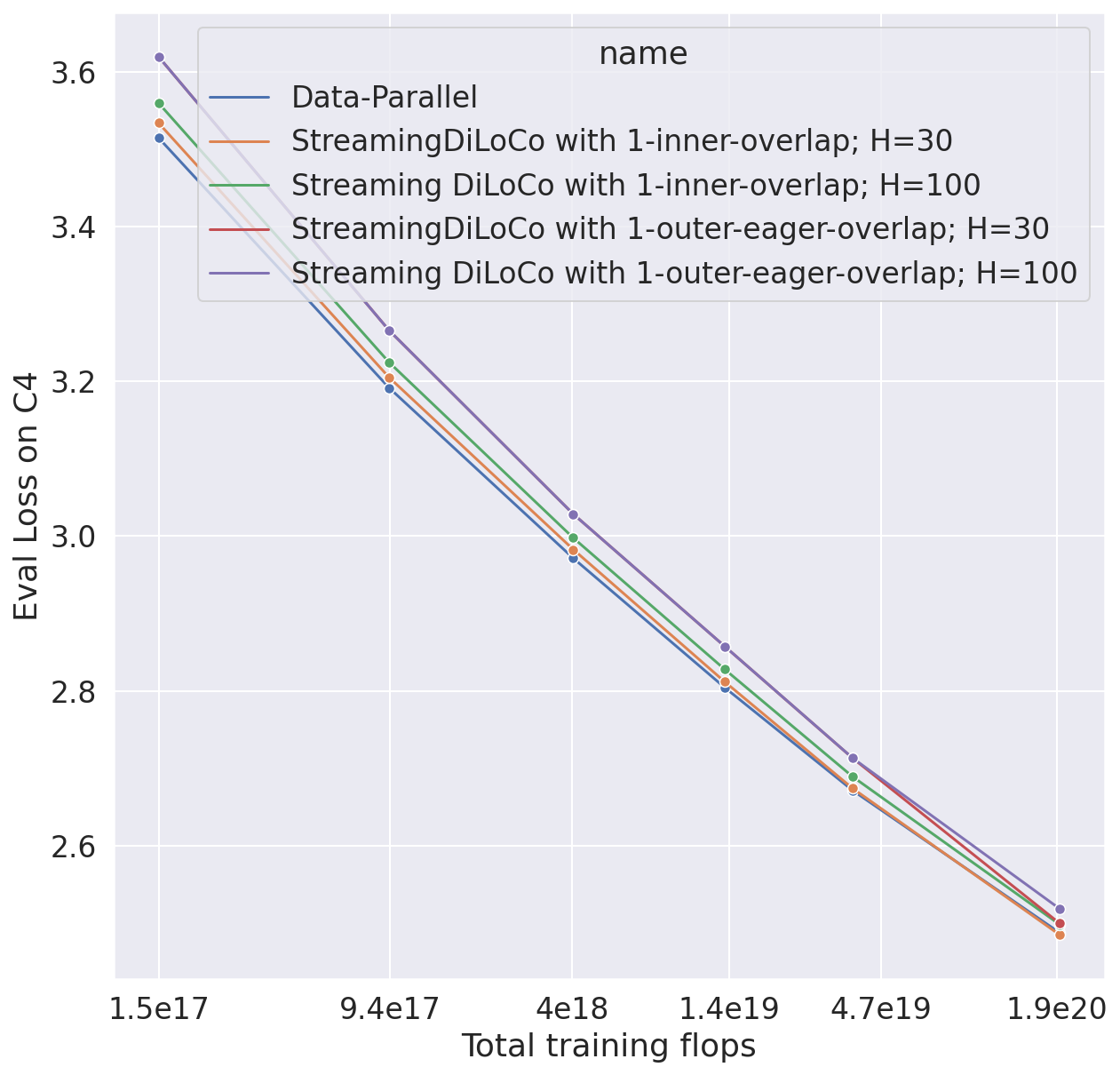}
  \caption{Evaluation loss on C4}
  \label{fig:scaling_loss}
\end{subfigure}\hfill
\begin{subfigure}{0.45\linewidth}
  \centering
  \includegraphics[width=1\linewidth]{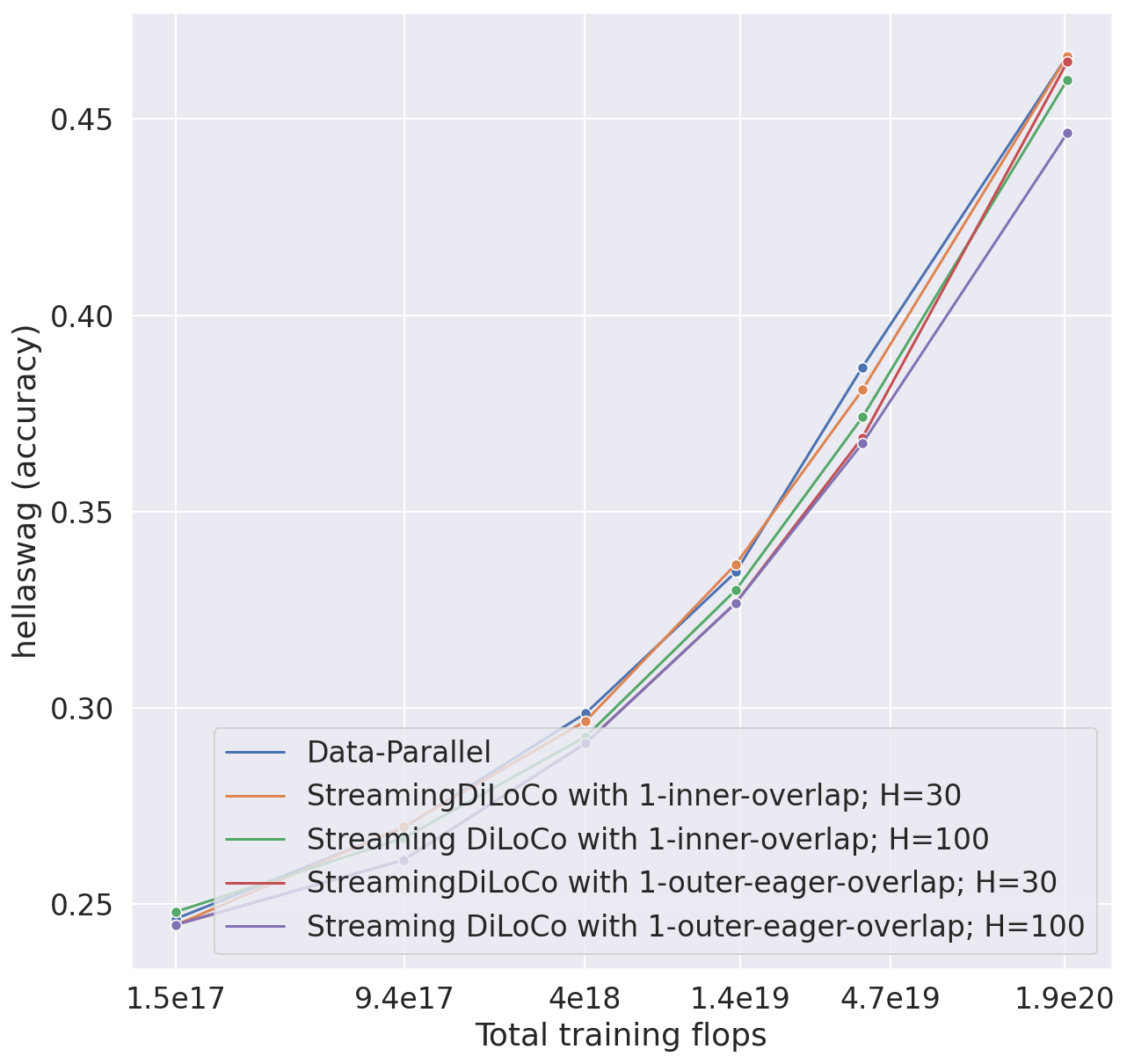}
  \caption{HellaSwag accuracy}
  \label{fig:scaling_hellaswag}
\end{subfigure}
\caption{\textbf{Scaling} models from 35M (1.49e17 flops) to 1B parameters (1.9e20 flops) on C4.}
\label{fig:scaling}
\end{figure*}

For training we use a modified version of the Nanodo codebase \citep{nanodo} that uses DrJax \citep{rush2024drjax} to parallelize inner steps across replicas. The inner optimization is done with an annotated variant of \texttt{jax.vmap} for the optimization step, with parameters having an extra leading axis for the DiLoCo replicas. The outer optimization is implemented with an all-reduce, without any central parameter server.

\subsection{Compute utilization simulation}

First, we simulate the training of a model using a DAG made of forward and backward nodes (refer to \cite{douillard2025streamingdiloco} for full details). We consider models of 1, 10, and 100 billion parameters with respectively a step time (pure compute) of 0.1, 0.8, and 4.9 seconds. For each model, we sweep a range of bandwidth from $10^{-1}$ Gbits/s to $10^{3}$ Gbits/s, and across different distributed training methods. We display the results of those simulation in \autoref{fig:bandwdith}.

As also noted by \cite{douillard2025streamingdiloco}, overlapping communication massively reduces required bandwidth, particularly as the models get larger and thus spend more time during computation. Indeed, as also shown in \autoref{tab:simulation} in the appendix, our method with 1-outer-step eager requires $1{,}177\times$ (471.5 vs 0.4) less Gbits/s than data-parallel for a 100 billion parameters model. Overlapping a single inner step, as proposed by \cite{douillard2025streamingdiloco}, only reduces required bandwidth by $336\times$ (471.5 vs 1.4).

\subsection{Scaling}

We display in \autoref{fig:scaling} the loss on C4 and the accuracy on HellaSwag \citep{zellers2019hellaswagmachinereallyfinish} of our model vs baselines from 35 million parameters to 1 billion parameters. We also report the full results, including accuracy on Piqa \citep{bisk2019piqareasoningphysicalcommonsense} and Arc-easy \citep{clark2018arc}, in \autoref{tab:scaling} in the appendix.

Notably, our method with 1-outer-step \textit{eager} (see \autoref{alg:eager-delayed}) with $H=30$ inner steps reaches the same performance as Data-Parallel at 1 billion scale, proving that our distributed method gets better at larger scale, where also the sheer size of the models make distributed methods ever more important. 

\begin{table*}[t]
\centering
\resizebox{1.0\linewidth}{!}{%
\begin{tabular}{@{}l|cccc|cccc@{}}
\toprule
Method & Token Budget & Hours spent w/ $+\infty$ Gbits/s & Hours spent w/ 1 Gbits/s & Terabytes exchanged & Eval Loss $\downarrow$ & HellaSwag $\uparrow$ & Piqa $\uparrow$ & Arc Easy $\uparrow$ \\
\midrule
\multirow{3}{*}{Data-Parallel} & 25B & 0.67 & 109 & 441 & 2.67 & \textbf{42.09} & 67.35 & \textbf{40.42} \\
 & 100B & 2.7 & 438 & 1,767 & 2.52 & 49.78 & 69.15 & \textbf{44.03} \\
 & 250B & 6.75 & 1097 & 4,418 & \textbf{2.45} & 53.86 & 70.45 & \textbf{44.21} \\
 \midrule
\multirow{3}{*}{\parbox{3.5cm}{\centering Streaming DiLoCo \\ with 1-inner-step overlap}} & 25B & 0.67 & 0.88 & 1.10 & \textbf{2.66} & 42.08 & \textbf{67.46} & 38.42 \\ 
& 100B & 2.7 & 3.5 & 4.42 & \textbf{2.51} & \textbf{49.98} & \textbf{69.96} & \textbf{44.03} \\
& 250B & 6.75 & 8.75 & 11.05 & \textbf{2.45} & \textbf{54.24} & \textbf{71.38} & 41.92 \\
\midrule
\multirow{3}{*}{\parbox{3.5cm}{\centering Streaming DiLoCo \\ with 1-outer-step overlap}} & 25B & 0.67 & 0.67 & 1.10 & 2.69 & 40.51 & 66.87 & 39.12 \\ 
& 100B & 2.7 & 2.7 & 4.42 & 2.53 & 49.48 & 68.82 & 41.05 \\
& 250B & 6.75 & 6.75 & 11.05 & 2.46 & 53.30 & 69.00 & 41.93 \\
\bottomrule
\end{tabular}
}
\caption{\textbf{Overtraining} Data-Parallel and our method on Dolma with a 1 billion parameters model. The latter performs slightly better despite exchanging in total $400\times$ fewer bits, reducing the peak bandwidth by $8\times$, and with a significantly relaxed training communication latency constraint: allowing communication to be as long as a full inner optimization phase.}
\label{tab:overtrain_steps}
\end{table*}

\subsection{Overtraining on Dolma}

Previous experiments on C4 are done with a token budget "optimal" according to Chinchilla scaling laws. However, nowadays LLMs are usually \textit{overtrained} with a significantly larger budget \citep{wortsman2023smallscaleproxieslargescaletransformer}, leading to better performance. Therefore, following \cite{douillard2025streamingdiloco}, we consider, for a 1 billion parameters model, three token budgets on the Dolma dataset \citep{soldaini2024dolmao}: 25, 100, and 250 billion tokens, which are respectively $1\times$, $4\times$, and $10\times$ larger than the ``optimal''. We display the results of our 1-outer-step eager method in \autoref{tab:overtrain_steps} alongside a data-parallel baseline and the techniques in \cite{douillard2025streamingdiloco} with 1-inner-step overlap. Two things are to be noted: (1) with a minimal 1 Gbits/s bandwidth, our model has close to 100\% compute utilization, while data-parallel training suffers massively, and (2) the performance of our method is lower than the less bandwidth efficient 1-inner-step overlapping, but we see the performance gap reduces as the token budget increases. We see that last observation as a hopeful perspective for distributed methods like the one in this paper, in a world where massively overtrained models are becoming the norm.

\subsection{Ablations}

We perform in this section ablations of our proposed method. All experiments are conducted on a 500 million parameters model trained on the C4 dataset.

\begin{figure}[t]
\centering
  \includegraphics[width=1\linewidth]{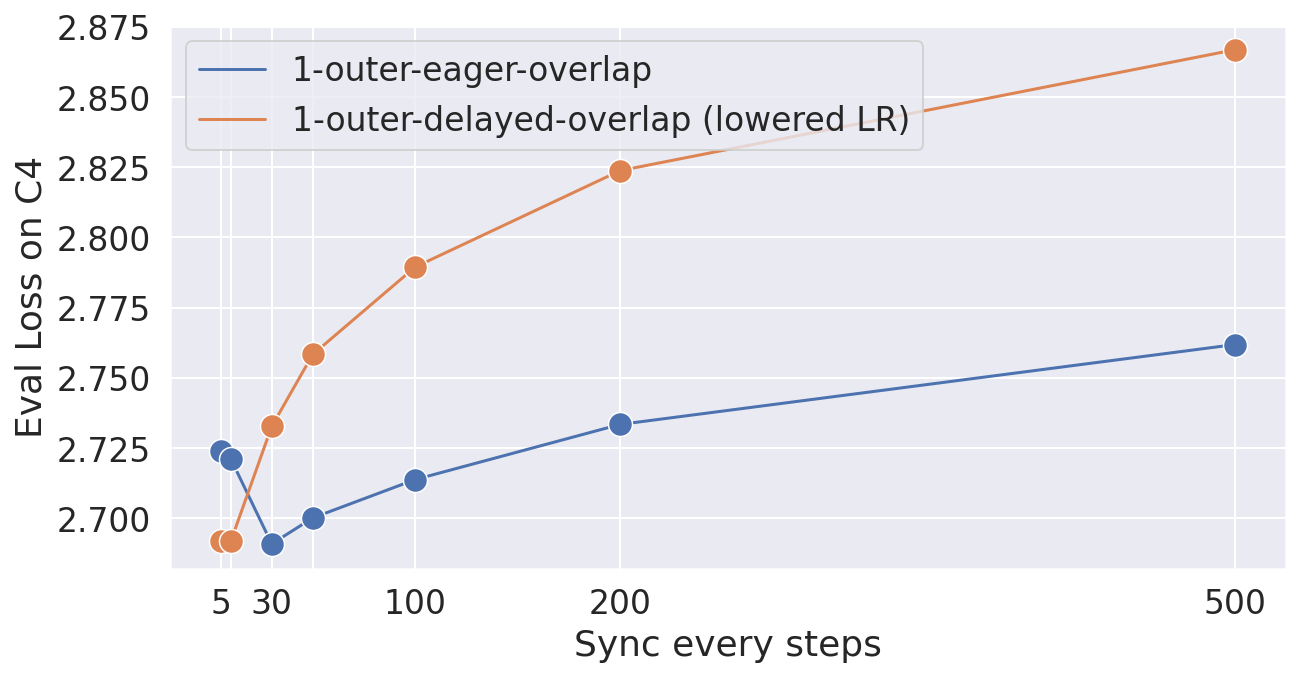}
  \caption{\textbf{Comparison of overlapping communication} over an outer step, using the na\"ive delayed version (\autoref{alg:na\"ive-delayed}) and the eager version (\autoref{alg:eager-delayed}) when varying the number of inner steps $H$.}
  \label{fig:stale_vs_eager}
\end{figure}

\paragraph{Number of inner steps.} We compare in \autoref{fig:stale_vs_eager} the loss on C4 of 1-outer-step na\"ive delayed (\autoref{alg:na\"ive-delayed}) vs 1-outer-step eager (\autoref{alg:eager-delayed}) while varying the number of inner steps $H$. An outer step is run after every $H$ inner steps, therefore our outer-step overlapping method will have increased amount of time to hide communication as $H$ increases, in addition of synchronizing less often. While for the eager method we keep the same hyperparameters as proposed by \cite{fig:streaming_diloco}, for the na\"ive delayed method we have to lower the outer learning rate by $4\times$ for stability reasons. Despite, we see that the na\"ive delayed version (in \textcolor{orange}{orange}) see a significant increase of loss as $H$ grows, while our eager version (in \textcolor{blue}{blue}) stays relatively constant. Na\"ive delayed sees an increase of loss of 6\% from $H=5$ vs $H=500$ while eager sees an increase of only 2\% from $H=30$ vs $H=500$.

\begin{figure}[t]
\centering
  \includegraphics[width=1\linewidth]{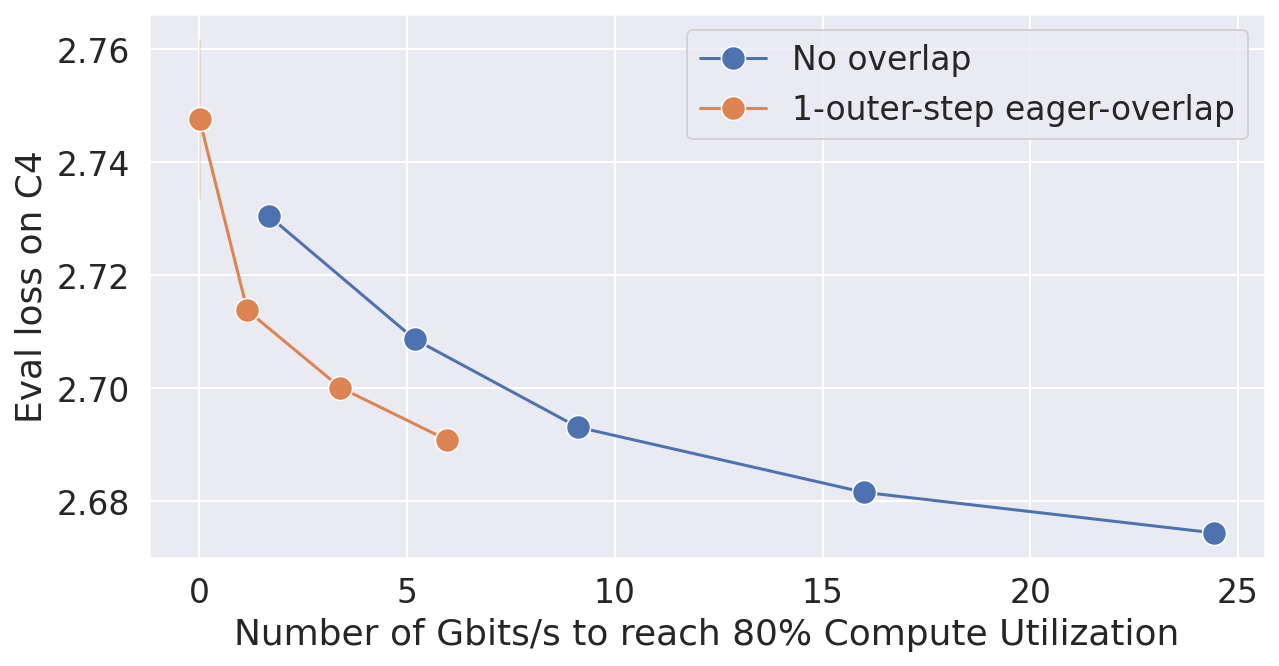}
  \caption{Varying the number of \textbf{inner steps}, which affects both the loss and the bandwidth required to reach a certain level of compute utilization. When bandwidth is scarce, it is preferable to overlap communication across an outer step.}
  \label{fig:cu_vs_loss}
\end{figure}

\paragraph{Loss vs bandwidth} The main application of our method is to massively reduce the required bandwidth. Therefore, we display in \autoref{fig:cu_vs_loss} a number of Gbit/s seconds required to reach a compute utilization of 80\% (i.e. only 20\% of the time is spent in communication) of Streaming DiLoCo without communication overlap (in \textcolor{blue}{blue}) vs 1-outer-step overlap with our eager method (in \textcolor{orange}{orange}). We vary for both methods the number of inner steps $H$. While no overlap can reach lower loss, it also requires more bandwidth, and in a particular constrained setting, our method can strike a better tradeoff. This is even more true for larger models, where the step time is longer, and thus we have more time to overlap communication.

\begin{figure}[t]
\centering
  \includegraphics[width=1\linewidth]{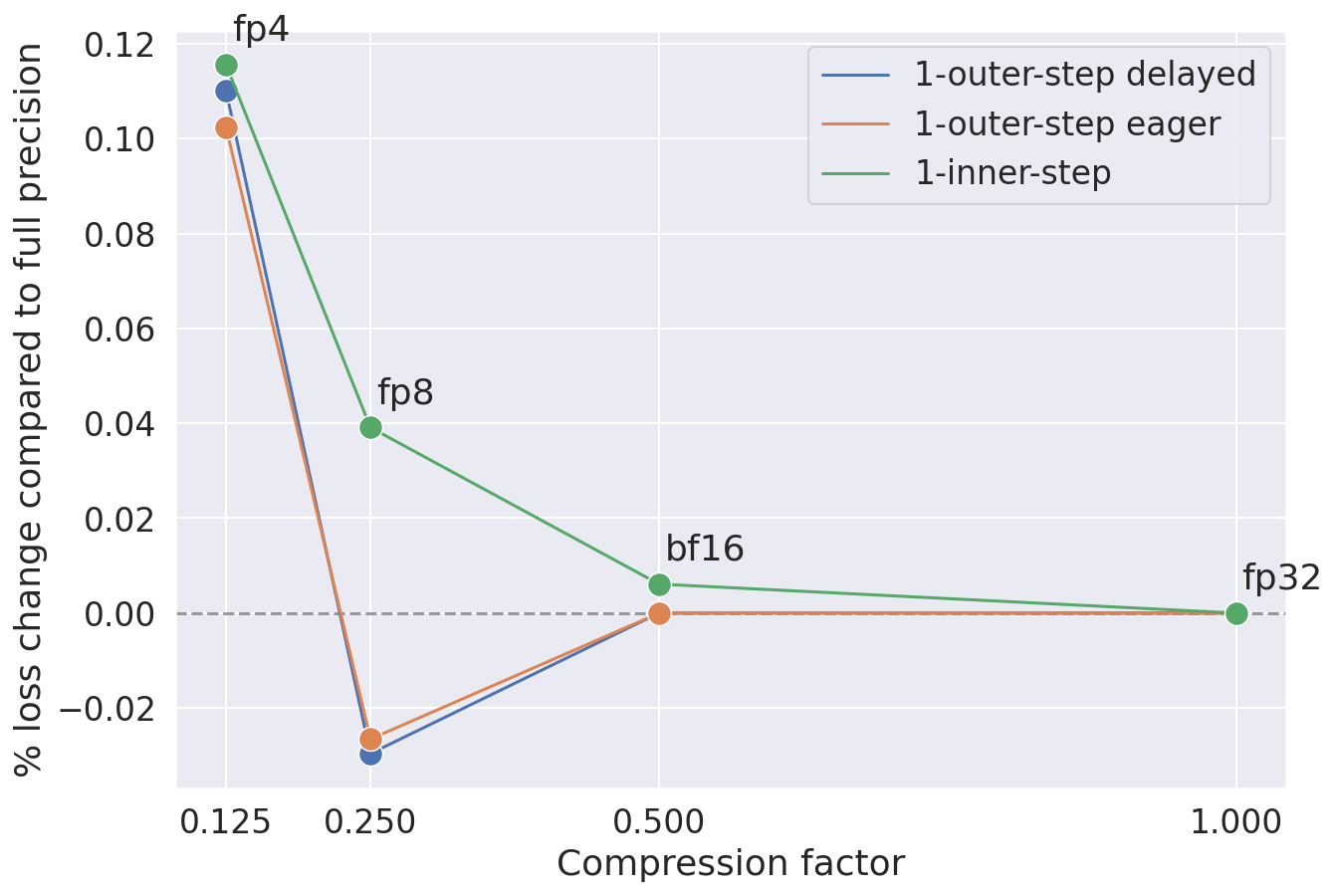}
  \caption{\textbf{Quantized communication} across three overlapping communication schemes: 1) 1-inner-step from \citep{douillard2025streamingdiloco}, 2) 1-outer-step na\"ive delayed from \autoref{alg:na\"ive-delayed}, and 3) 1-outer-step eager from \autoref{alg:eager-delayed}.} 
  \label{fig:compression}
\end{figure}

\paragraph{Quantized Communication} Streaming DiLoCo \citep{douillard2025streamingdiloco} proposed to quantize the outer gradients to lower precision, and consider float32 and bfloat16, and float8 and float4 using EXMY \citep{agrawal2024exmy}. We consider whether our outer overlapping communication can have negative interaction with quantized communication, and ablates the percentage loss of difference v.s. using using full precision in \autoref{fig:compression}. Notably, as \cite{douillard2025streamingdiloco}'s inner-step overlap, the difference stays bounded by 0.12\%, which is minimal considered the reduction of bandwidth provided.

\begin{table*}[t]
\centering
\resizebox{1.0\linewidth}{!}{%
\begin{tabular}{@{}l|c|cccc@{}}
\toprule
Communication overlap & Tolerated latency in sec. $\uparrow$ & Eval Loss $\downarrow$ & HellaSwag $\uparrow$ & Piqa $\uparrow$ & Arc-Easy $\uparrow$ \\
\midrule
No overlap & 0 & \textbf{2.67} & \textbf{38.26} & 66.59 & 34.91 \\
1-inner-step & 0.08 & \textbf{2.67} & 37.96 & 66.10 & \textbf{36.14} \\
1-outer-step delayed & 2.4 & 3.01 & 29.40 & 60.93 & 34.73 \\
1-outer-step delayed, lowered LR  & 2.4& 2.73 & 35.83 & 64.96 & 34.21 \\
1-outer-step eager &  2.4 & 2.69 & 37.52 & \textbf{66.86} & 34.91 \\
2-outer-steps eager & \textbf{4.8} & 2.73 & 36.47 & 64.85 & 35.43 \\
\bottomrule
\end{tabular}
}
\caption{\textbf{Communication overlap} comparison for a 500M parameters model, performing a step (forward \& backward) in 0.08 seconds. Overlapping 1-inner-step as proposed by \citep{douillard2025streamingdiloco} allows communication to take 0.08 seconds, while we propose to overlap up to 2.4 seconds ($H=30$ total steps).}
\label{tab:com_overlap}
\end{table*}

\begin{table}[t]
\centering
\resizebox{1.0\linewidth}{!}{%
\begin{tabular}{@{}l|ll@{}}
\toprule
Overlapping & DiLoCo variant & Evaluation loss \\
\midrule
\multirow{2}{*}{No overlap} & DiLoCo & $2.68$ \\
& Streaming DiLoCo & $2.67_{\textcolor{ForestGreen}{-0.3\%}}$ \\
\multirow{2}{*}{1-outer-step eager} & DiLoCo & 2.69 \\
& Streaming DiLoCo & $2.71_{\textcolor{BrickRed}{+0.7\%}}$ \\
\bottomrule
\end{tabular}
}
\caption{\textbf{DiLoCo variant comparison} for no communication overlapping v.s. our 1-outer-step eager overlapping when varying the underlying DiLoCo algorithms: either the standard DiLoCo \citep{douillard2023diloco} where all parameters are synchronized together, or its streaming variant \citep{douillard2025streamingdiloco} with partial synchronization.}
\label{tab:vanilla}
\end{table}

\paragraph{Two-outer-steps overlap} While we perform all previous experiments with 1-outer-step overlap, whose total overlap length scales with the number of inner steps $H$, we also consider in \autoref{tab:com_overlap} doing 2-outer-steps eager overlap; thus overlapping for $2H$ inner steps. Notably, doing more than 1-outer-step overlap can be harmful (2.73 loss v.s. 2.67 when not doing any overlap, a 2.2\% increase). However, this difference may not always translate on downstream tasks, 2-outer-steps eager improves accuracy on Arc-Easy \citep{clark2018arc} by 1.4\%. Furthermore, the tolerated communication latency, how long communication is overlapped with computation, increases significantly to 4.8 seconds.

\paragraph{DiLoCo variants.} We consider the impact of the underlying DiLoCo variants choosen in \autoref{tab:vanilla}: either the standard DiLoCo \citep{douillard2023diloco} where all parameters are synchronized together, or its streaming variant \citep{douillard2025streamingdiloco} with partial synchronization. For both we compare no overlapping of communication v.s. the overlapping scheme proposed in \autoref{alg:eager-delayed}. We found a slight degradation of performance for 1-outer-step eager when using the streaming variant, however it is very limited (<1\% different in evaluation loss) and the bandwidth advantage brought by this variant outweighs the cost.

\section{Related Works}\label{sec:related_works}

\paragraph{Federated learning / local SGD.}  Differing from model merging's single combination step, Federated Averaging (FedAvg) \citep{mcmahan2017fedavg} and Local SGD \citep{stich2019local} iteratively combine models with the goal of minimizing bandwidth requirements. They operate by performing local training, typically using SGD, across workers for a certain number of steps before implementing some form of worker parameter synchronization or parameter aggregation.  In their original formulations, both FedAvg and Local SGD employed a straightforward average of parameters across workers.  As demonstrated by \cite{reddi2021adaptive}, synchronization becomes more effective when each worker computes a ``model delta,'' which are then aggregated to produce a pseudo-gradient, also termed an \textit{outer gradient}, subsequently utilized by a first-order optimizer ~\citep{reddi2021adaptive,ilharco2022patching}.  This yields a bi-level optimization framework with inner optimizers and an outer optimizer, referred to as FedOpt by \cite{reddi2021adaptive}, who propose using SGD as the inner optimizer and adaptive techniques such as Adam~\citep{kingma2014adam} as the outer optimizer in resource-constrained Federated Learning settings.

\paragraph{Distributed training for LLMs.} The increasing computational demands of training large language models (LLMs) have accelerated the need for distributed methodologies, applicable to both inference \citep{borzunov2023petals} and training \citep{presser2020stub,diskin2021distributedcollab,ryabinin2021moshpit}. More recently, DiLoCo \citep{douillard2023diloco} introduced a specific instantiation of FedOpt \citep{reddi2021adaptive} utilizing AdamW \citep{loshchilov2018adamw} as the inner optimizer and Nesterov \citep{sutskever2013nesterov} as the outer optimizer \citep{huo2020outernesterov}. This simple formulation has proven effective for distributed training with LLMs, particularly in scenarios with a limited number of replicas (under 100) and without replica sampling, aligning more closely with cross-silo federated learning \citep{kairouz2021advances}. The FedOpt algorithm has also been demonstrated to be effective in training LLMs in settings that resemble cross-device federated learning~\citep{charles2024towards}. The empirical effectiveness of DiLoCo has been reproduced in multiple studies \citep{jaghouar2024opendiloco,sani2024futurelargelanguagemodel} and successfully scaled to models with 10 billion parameters \citep{jaghouar2024intellect1}.  In related research, a minor modification to the way the outer Nesterov accumulates outer gradients has shown improved handling of asynchronicity among workers with different processing speeds \citep{liu2024asyncdiloco}.  DiLoCo provides an additional axis of parallelism to distributed training \citep{shoeybi2020megatronlmtrainingmultibillionparameter} and is compatible \citep{jaghouar2024intellect1} with other existing parallelization approaches such as FSDP \citep{zhao2023fsdp}, or even another layer of federated learning \citep{sani2024photonfederatedllmpretraining}. More recently, \cite{douillard2025streamingdiloco} proposed Streaming DiLoCo where only a subset of the parameters are shared at each given synchronization round, in effect lowering the peak required bandwidth.

\paragraph{Overlapping Communication.} Overlapping communication with computation is critical in many aspects of distributed training in order to minimize the time waiting for communication, and thus maximizing computation. Methods have been designed to minimize the ``bubble-of-time'' in pipeline parallelism \citep{narayanan2021pipedream2bw,qi2023zerobubblepipelineparallelism}, in data-parallel \citep{zhao2013butterfly,lin2020deepgradientcompressionreducing}, and federated learning \citep{xie2019asynchronous,liu2024asyncdiloco}. In particular, in the latter case of federated learning, it is particularly useful to handle the case of ``stragglers'' where some replicas are slower than others \citep{koh2006parallel, recht2011hogwild, dean2012large, lian2015asynchronous, diskin2021distributedcollab}.

\section{Conclusion}\label{sec:conclusion}

In this work, we present an improvement over DiLoCo, allowing us to overlap the communication of the ``outer gradients'' by a whole synchronization round (an outer step), which can be as much as hundreds of inner optimization steps. We show that a na\"ive formulation results in worse performance, particularly with untuned hyperparameters, and propose instead an ``eager'' variant. In this variant, we decouple the per-replica local outer gradients with the synchronized, and thus delayed, outer gradients. Indeed, we apply a mixture of the current local outer gradient with the delayed mixture of all other replicas' outer gradients. This improved formulation of communication overlapping results in minimal performance degradation, particularly when the model is trained for a large token budget, as is currently the norm, and when the model scale is large. The success of the eager technique in overlapped communication and computation suggests further applications in distributed optimization, e.g. overlapping only few inner steps rather an entire inner optimization phase, which can be useful in settings where bandwidth is not as constrained. Another interesting direction for further work is developing a convergence theory for delayed outer gradients.


\clearpage
\section*{Acknowledgements}

We thank Arthur Szlam and Marc'Aurelio Ranzato for advices and general support. We also thank Gabriel Teston, Zachary Charles, Zachary Garrett, and Keith Rush for providing engineering support. Finally, we thank Jeff Dean, Raia Hadsell, and Koray Kavukcuoglu for leadership support.

\bibliographystyle{plainnat}
\nobibliography*
\bibliography{main}

%

\clearpage
\section*{Supplementary Materials} \label{sec:supp}

\begin{table}[!ht]
\centering
\resizebox{1.0\linewidth}{!}{%
\begin{tabular}{@{}l|cccc@{}}
\toprule
Model scale & Hidden dim & Num layers & Num heads & Token budget \\
\midrule
35M & $2{,}048$ & 6 & 8 & 700M\\
100M & $3{,}072$ & 9 & 12 & 1.5B \\
200M & $4{,}096$ & 12 & 16 & 3.5B \\
300M & $5{,}120$ & 15 & 20 & 6B\\
500M & $6{,}144$ & 18 & 24 & 11B \\
1B & $8{,}192$ & 24 & 32 & 25B\\
\bottomrule
\end{tabular}
}
\caption{\textbf{Architecture hyperparameters}: we consider model from 35M to 1B with the following hyperameters and chinchilla-optimal token budget. For all model scale, the vocabulary size is $32{,}000$.}
\label{tab:hp_architecture}
\end{table}

\begin{table*}[t]
\centering
\resizebox{1.0\linewidth}{!}{%
\begin{tabular}{@{}ccc|c|ccccc@{}}
\toprule
\multirow{2}{*}{Model size} & \multirow{2}{*}{\# layers} & \multirow{2}{*}{Step time} &   \multirow{2}{*}{Method} & \multicolumn{5}{c}{Gbit/s to reach a compute utilization $\texttt{CU} = $?} \\
&&&&$50\%$ &  $80\%$ &  $90\%$ &  $95\%$ &  $99\%$  \\
\midrule
\multirow{5}{*}{1B} & \multirow{5}{*}{24} & \multirow{5}{*}{0.1s} & Data-Parallel & 86.8 & 152.6 & 184.2 & 222.3 & 569.0 \\
& & & Streaming DiLoCo & 1.4 & 5.2 & 9.1 & 16.0 & 28.1 \\
& & & Streaming DiLoCo with 1-inner-step overlapped FP4 com. H=30 & 2.4 & 3.6 & 4.3 & 4.3 & 4.3 \\
& & & Streaming DiLoCo with 1-inner-step overlapped FP4 com. H=100 & 0.4 & 0.9 & 1.7 & 2.0 & 3.0 \\
& & & Streaming DiLoCo with 1-outer-step overlapped FP4 com. H=30 & 1.1 & 2.0 & 2.0 & 2.0 & 2.4 \\
& & & Streaming DiLoCo with 1-outer-step overlapped FP4 com. H=100 & 0.1 & 0.2 & 0.2 & 0.2 & 0.2 \\
\midrule
\multirow{5}{*}{10B} & \multirow{5}{*}{48} & \multirow{5}{*}{0.8s} & Data-Parallel & 104.8 & 222.3 & 222.3 & 268.3 & 471.5 \\
& & & Streaming DiLoCo & 1.7 & 5.2 & 9.1 & 13.3 & 19.3 \\
& & & Streaming DiLoCo with 1-inner-step overlapped FP4 com. H=30 & 1.4 & 2.4 & 2.4 & 3.0 & 3.0 \\
& & & Streaming DiLoCo with 1-inner-step overlapped FP4 com. H=100 & 0.4 & 0.9 & 1.4 & 1.4 & 1.7 \\
& & & Streaming DiLoCo with 1-outer-step overlapped FP4 com. H=30 & 0.7 & 1.1 & 1.4 & 1.4 & 1.4 \\
& & & Streaming DiLoCo with 1-outer-step overlapped FP4 com. H=100 & 0.1 & 0.2 & 0.2 & 0.3 & 0.3 \\
\midrule
\multirow{5}{*}{100B} & \multirow{5}{*}{108} & \multirow{5}{*}{4.9s} & Data-Parallel & 184.2 & 323.8 & 390.7 & 390.7 & 471.5 \\
& & & Streaming DiLoCo & 2.4 & 6.2 & 9.1 & 11.0 & 19.3 \\
& & & Streaming DiLoCo with 1-inner-step overlapped FP4 com. H=30 & 0.9 & 1.7 & 2.0 & 2.0 & 2.0 \\
& & & Streaming DiLoCo with 1-inner-step overlapped FP4 com. H=100 & 0.5 & 0.9 & 1.1 & 1.1 & 1.4 \\
& & & Streaming DiLoCo with 1-outer-step overlapped FP4 com. H=30 & 0.5 & 0.8 & 0.9 & 0.9 & 0.9 \\
& & & Streaming DiLoCo with 1-outer-step overlapped FP4 com. H=100 & 0.2 & 0.3 & 0.3 & 0.3 & 0.4 \\
\bottomrule
\end{tabular}
}
\caption{\textbf{Simulation}: we estimate the step time (pure compute) of 10B and 100B based on the required flops using \cite{kaplan2020scalinglawsneurallanguage} rule and using a MFU of 60\%. For all DiLoCo and Streaming DiLoCo-variants, we use $H=100$. For all Streaming DiLoCo-variants, we use a fragment size of 3 layers.}
\label{tab:simulation}
\end{table*}

\begin{table*}[t]
\centering
\resizebox{1.0\linewidth}{!}{%
\begin{tabular}{@{}cc|ccc|cccc@{}}
\toprule
Model size & Flops & Method & $H$ & \# overlapped steps &  Eval Loss $\downarrow$ & HellaSwag $\uparrow$ & Piqa $\uparrow$ & Arc Easy $\uparrow$ \\
\midrule
\multirow{6}{*}{35M} & \multirow{6}{*}{1.5e17} & Data-Parallel  & 0 & 0 & 3.51 & 24.62 & 57.89 & 29.65 \\
& &  DiLoCo   & 30 & 0 & 3.54 & 24.53 & 58.11 & 29.65 \\
& &  Streaming DiLoCo with 1-inner-overlap   & 30 & 1 & 3.53 & 24.46 & 57.67 & 30.53 \\
& &  Streaming DiLoCo with 1-inner-overlap   & 100 & 1 & 3.56 & 24.80 & 57.89 & 29.12 \\
& &  Streaming DiLoCo with 1-outer-eager-overlap   & 30 & 30 & 3.62 & 24.47 & 56.58 & 27.19 \\
& &  Streaming DiLoCo with 1-outer-eager-overlap   & 100 & 100 & 3.62 & 24.47 & 56.58 & 27.19 \\
\midrule
\multirow{6}{*}{100M} & \multirow{6}{*}{9.4e17} & Data-Parallel  & 0 & 0 & 3.19 & 26.94 & 60.12 & 30.35 \\
& &  DiLoCo   & 30 & 0 & 3.21 & 26.59 & 60.50 & 29.12 \\
& &  Streaming DiLoCo with 1-inner-overlap   & 30 & 1 & 3.21 & 26.97 & 59.58 & 31.40 \\
& &  Streaming DiLoCo with 1-inner-overlap   & 100 & 1 & 3.22 & 26.68 & 60.39 & 31.93 \\
& &  Streaming DiLoCo with 1-outer-eager-overlap   & 30 & 30 & 3.27 & 26.12 & 59.19 & 28.77 \\
& &  Streaming DiLoCo with 1-outer-eager-overlap   & 100 & 100 & 3.27 & 26.12 & 59.19 & 28.77 \\
\midrule
\multirow{6}{*}{200M} & \multirow{6}{*}{4e18} & Data-Parallel  & 0 & 0 & 2.97 & 29.86 & 63.71 & 35.44 \\
& &  DiLoCo   & 30 & 0 & 2.98 & 29.71 & 62.30 & 33.68 \\
& &  Streaming DiLoCo with 1-inner-overlap   & 30 & 1 & 2.98 & 29.67 & 61.92 & 34.39 \\
& &  Streaming DiLoCo with 1-inner-overlap   & 100 & 1 & 3.00 & 29.27 & 62.13 & 34.21 \\
& &  Streaming DiLoCo with 1-outer-eager-overlap   & 30 & 30 & 3.03 & 29.10 & 61.70 & 32.81 \\
& &  Streaming DiLoCo with 1-outer-eager-overlap   & 100 & 100 & 3.03 & 29.10 & 61.70 & 32.81 \\
\midrule
\multirow{6}{*}{300M} & \multirow{6}{*}{1.4e19} & Data-Parallel  & 0 & 0 & 2.80 & 33.46 & 64.69 & 34.91 \\
& &  DiLoCo   & 30 & 0 & 2.81 & 33.87 & 64.74 & 34.74 \\
& &  Streaming DiLoCo with 1-inner-overlap   & 30 & 1 & 2.81 & 33.66 & 63.49 & 35.09 \\
& &  Streaming DiLoCo with 1-inner-overlap   & 100 & 1 & 2.83 & 33.00 & 63.71 & 34.39 \\
& &  Streaming DiLoCo with 1-outer-eager-overlap   & 30 & 30 & 2.86 & 32.67 & 65.34 & 35.44 \\
& &  Streaming DiLoCo with 1-outer-eager-overlap   & 100 & 100 & 2.86 & 32.67 & 65.34 & 35.44 \\
\midrule
\multirow{6}{*}{500M} & \multirow{6}{*}{4.7e19} & Data-Parallel  & 0 & 0 & 2.67 & 38.68 & 66.49 & 37.19 \\
& &  DiLoCo   & 30 & 0 & 2.68 & 38.37 & 65.61 & 36.32 \\
& &  Streaming DiLoCo with 1-inner-overlap   & 30 & 1 & 2.67 & 38.10 & 66.21 & 34.91 \\
& &  Streaming DiLoCo with 1-inner-overlap   & 100 & 1 & 2.69 & 37.40 & 65.51 & 34.74 \\
& &  Streaming DiLoCo with 1-outer-eager-overlap   & 30 & 30 & 2.71 & 36.89 & 65.61 & 35.44 \\
& &  Streaming DiLoCo with 1-outer-eager-overlap   & 100 & 100 & 2.71 & 36.74 & 65.56 & 35.79 \\
\midrule
\multirow{6}{*}{1B} & \multirow{6}{*}{1.9e20} & Data-Parallel  & 0 & 0 & 2.49 & 46.60 & 68.93 & 39.65 \\
& &  DiLoCo   & 30 & 0 & 2.49 & 46.56 & 68.82 & 36.84 \\
& &  Streaming DiLoCo with 1-inner-overlap   & 30 & 1 & 2.48 & 46.60 & 69.04 & 39.12 \\
& &  Streaming DiLoCo with 1-inner-overlap   & 100 & 1 & 2.50 & 46.00 & 68.82 & 38.42 \\
& &  Streaming DiLoCo with 1-outer-eager-overlap   & 30 & 30 & 2.50 & 46.45 & 68.50 & 39.47 \\
& &  Streaming DiLoCo with 1-outer-eager-overlap   & 100 & 100 & 2.52 & 44.64 & 68.12 & 36.14 \\
\bottomrule
\end{tabular}
}
\caption{\textbf{Scaling} from 35 million parameters to 4 billion parameters using a chinchilla-optimal number of flops/tokens. We train on the C4 dataset, and report the evaluation loss on its validation set.}
\label{tab:scaling}
\end{table*}

\end{document}